\def\arxivversion{1}
\documentclass[conference]{IEEEtran}
\IEEEoverridecommandlockouts

\usepackage[T1]{fontenc}
\usepackage{cite}
\usepackage{amsmath,amssymb,amsfonts}
\usepackage{algorithmic}
\usepackage{graphicx}
\usepackage{textcomp}
\usepackage{xcolor}
\usepackage{multirow}
\usepackage{subcaption}
\usepackage{booktabs}
\usepackage{float}
\usepackage{notoccite}
\def\BibTeX{{\rm B\kern-.05em{\sc i\kern-.025em b}\kern-.08em
    T\kern-.1667em\lower.7ex\hbox{E}\kern-.125emX}}

\begin{document}

\title{Event-Driven Language Models with Sparse Neural Activity for Neuromorphic Hardware}

\ifdefined\arxivversion
\newcommand{\arxivcopyrightnotice}{© 2026 IEEE. Personal use of this material is permitted. Permission from IEEE must be obtained for all other uses, in any current or future media, including reprinting/republishing this material for advertising or promotional purposes, creating new collective works, for resale or redistribution to servers or lists, or reuse of any copyrighted component of this work in other works.}
\fi

\author{\IEEEauthorblockN{Simon Richter$^{1}$, Ruhai Lin$^{2}$, Jason Yik$^{3}$, Taylor Kergan$^{2}$, Rui-Jie Zhu$^{2}$, Farshad Moradi$^{4}$, and Jason Eshraghian$^{2}$}
\IEEEauthorblockA{\textit{$^{1}$Department of Electrical and Computer Engineering, Aarhus University, Aarhus, Denmark}\\
\textit{$^{2}$Department of Computer Science and Engineering, University of California, Santa Cruz, Santa Cruz, USA}\\
\textit{$^{3}$School of Engineering and Applied Sciences, Harvard University, Cambridge, USA}\\
\textit{$^{4}$Department of Electrical and Computer Engineering, University of Southern Denmark, Odense, Denmark}\\
\texttt{\{siri, rlin50, tkergan, ridger, jsn\}}@ucsc.edu, \texttt{jyik@g.harvard.edu}, \texttt{moradi@sdu.dk}}
\ifdefined\arxivversion\thanks{\arxivcopyrightnotice}\fi
}

\maketitle

\begin{abstract}
Inference with transformer-based large language models (LLMs) is often limited by the memory-bound KV cache and quadratic attention cost. State-space models (SSMs) mitigate this through linear attention and fixed-size recurrent states, but their large dense linear projections remain computationally expensive even after quantization. We introduce a method that induces sparse neural activity in heavily quantized linear-attention models with minimal performance loss. Activations below a per-projection trainable threshold ($\pm \Delta$) are nullified while preserving crucial outliers, achieving comparable performance to dense models with up to 4$\times$ fewer effective arithmetic operations. Targeting a multi-core, multi-chip neuromorphic platform, where event-driven execution converts unstructured sparsity into throughput at both the compute and communication levels, a capability GPU architectures fundamentally lack, we project up to \textbf{37$\times$ higher throughput} and \textbf{16$\times$ lower power} versus edge GPU inference of a comparable transformer-based model, and up to \textbf{5.4$\times$ improvements} over the non-sparsified baseline. These results position sparse, quantized linear-attention models as a natural fit for deploying LLMs on event-driven multi-core platforms.
\end{abstract}


\section{Introduction}
The growing scale of Large Language Models (LLMs) presents significant challenges, driven primarily by the self-attention mechanism whose cost scales quadratically with sequence length. This bottleneck makes long-context applications prohibitively expensive in resource-constrained settings. Linear-attention models and State space models (SSMs) have emerged as a powerful alternative, replacing quadratic attention with a linear-time recurrent mechanism and achieving competitive results across diverse domains \cite{gu2024mambalineartimesequencemodeling, ssm_geonomics, wang2025systematic, voelker2019legendre}. Nevertheless, the billions of floating-point operations (FLOPs) required by SSMs still impede their deployment on edge devices, where low latency and energy efficiency are critical.


Previous model compression techniques have focused on reducing either the number of weights or activations, primarily through pruning. Weight pruning permanently removes parameters from the model, while activation pruning targets the intermediate outputs during inference. Pruning can be unstructured, removing individual weights or activations, or structured, which eliminates entire channels or blocks. While structured pruning is easier to exploit on modern specialized GPUs, the lack of fine-grain control often leads to lower model performance compared to unstructured pruning \cite{cheng2024surveydeepneuralnetwork}. Theoretically, unstructured activation sparsity is promising because zero-valued activations can dynamically eliminate entire rows from memory access and subsequent Matrix-Vector Multiplication (MVM) operations without arbitrary (i.e. block-wise or channel wise) constraints. However, realizing benefits from unstructured sparsity is challenging for three primary reasons: 1) In long sequences, the self-attention mechanism becomes the primary memory bottleneck, diminishing any performance gains from sparsity in the dense linear layers; 2) Current methods for inducing unstructured sparsity typically achieve only modest levels model-wide or else significantly degrade model performance, and 3) GPU memory interfaces (HBM) are optimized for contiguous data access and are ill-suited for the fine-grained, irregular memory patterns that result from unstructured sparsity.

SSMs inherently address the first challenge above by replacing the quadratic self-attention mechanism, thereby removing its associated memory bottleneck. In this work, we tackle the remaining two issues by introducing a method to induce high activation sparsity in quantized linear-attention models with minimal performance degradation. We achieve this by injecting a sparsity-inducing pre-activation gate before each layer that in the forward pass, pushes activations within a learnable range of $\pm \Delta$ towards zero, while in the backward pass, it maintains a smooth gradient flow for these near-zero activations, ensuring stable training.
Crucially, the gate preserves high-magnitude activations (outliers), both positive and negative, which are known to be vital for LLM performance~\cite{xiao2023smoothquant,raman2025rethinking}. This allows the model to maintain expressiveness without greatly disrupting the original activation distribution. This approach yields model-wide MAC operation reduction from sparse activations of up to 76\% with negligible impact on performance or additional training time. Multi-core, multi-chip systems with event-driven processing elements are uniquely positioned to exploit this unstructured sparsity at two levels of the system hierarchy simultaneously: within each core, zero-valued activations can be skipped locally, reducing intra-chip compute and memory traffic; across chips, event-driven interconnects transmit only non-zero activations, directly cutting inter-chip communication overhead. Deploying our sparse model on such a multi-core, multi-chip platform, the benefits become substantial. We project up to 37$\times$ reduction in latency and a 16$\times$ reduction in energy-per-token compared to a similarly sized transformer-based model on an edge GPU. Compared to a dense version of the same SSM on the same neuromorphic hardware, our method shows a 5.4$\times$ improvement in both metrics during the generate phase and 3.5$\times$ during prefill.

\begin{figure*}[t]
    \centering
    \begin{minipage}{0.85\textwidth}
        \centering
        
        \begin{subfigure}[b]{0.45\textwidth}
            \centering
            \includegraphics[width=\textwidth]{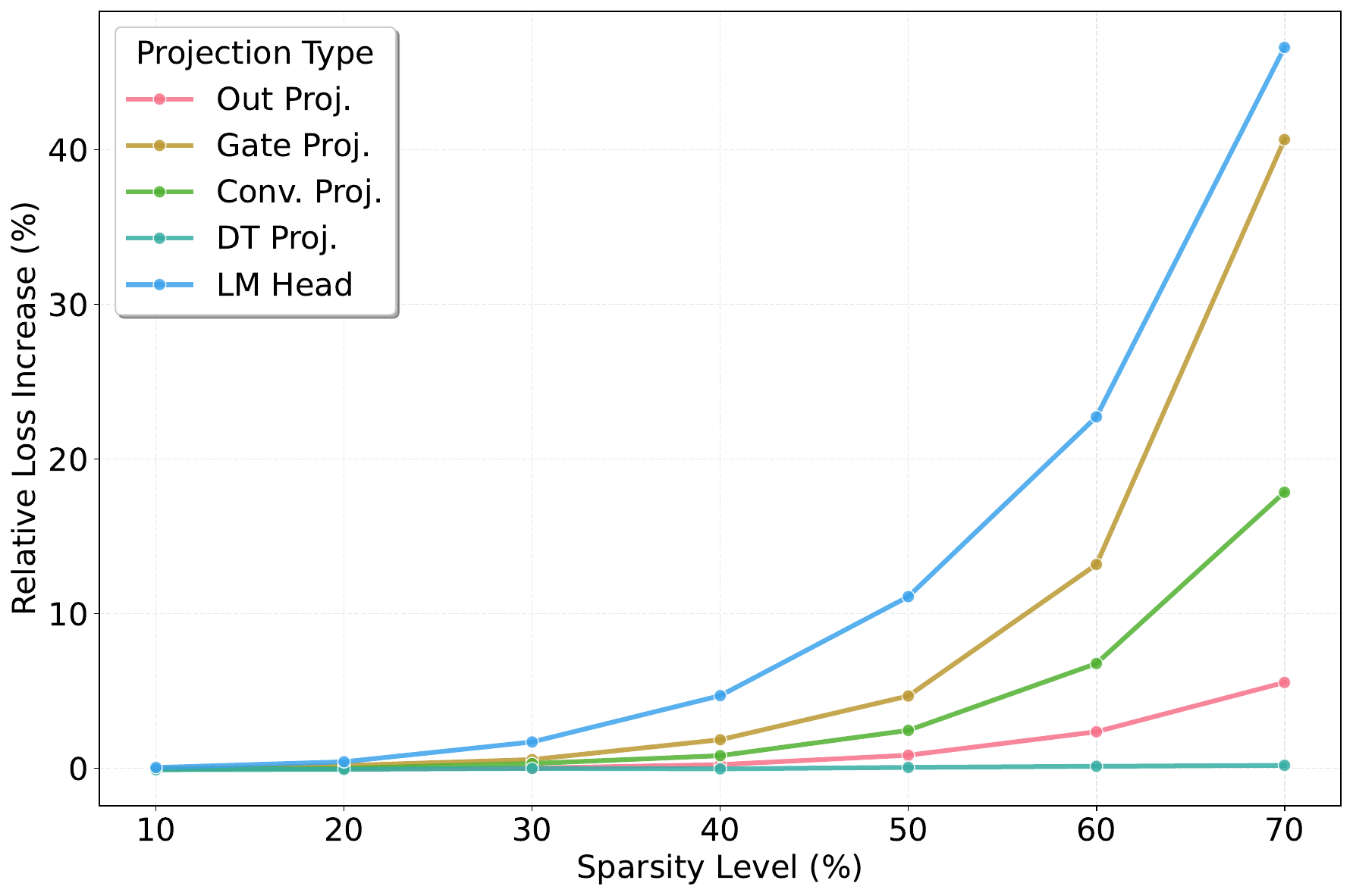}
            \caption{Projection-wise sensitivity to forced sparsity.}
            \label{fig:per_layer_sensitivity}
        \end{subfigure}
        \hspace{0.05\textwidth}
        \begin{subfigure}[b]{0.45\textwidth}
            \centering
            \includegraphics[width=\textwidth]{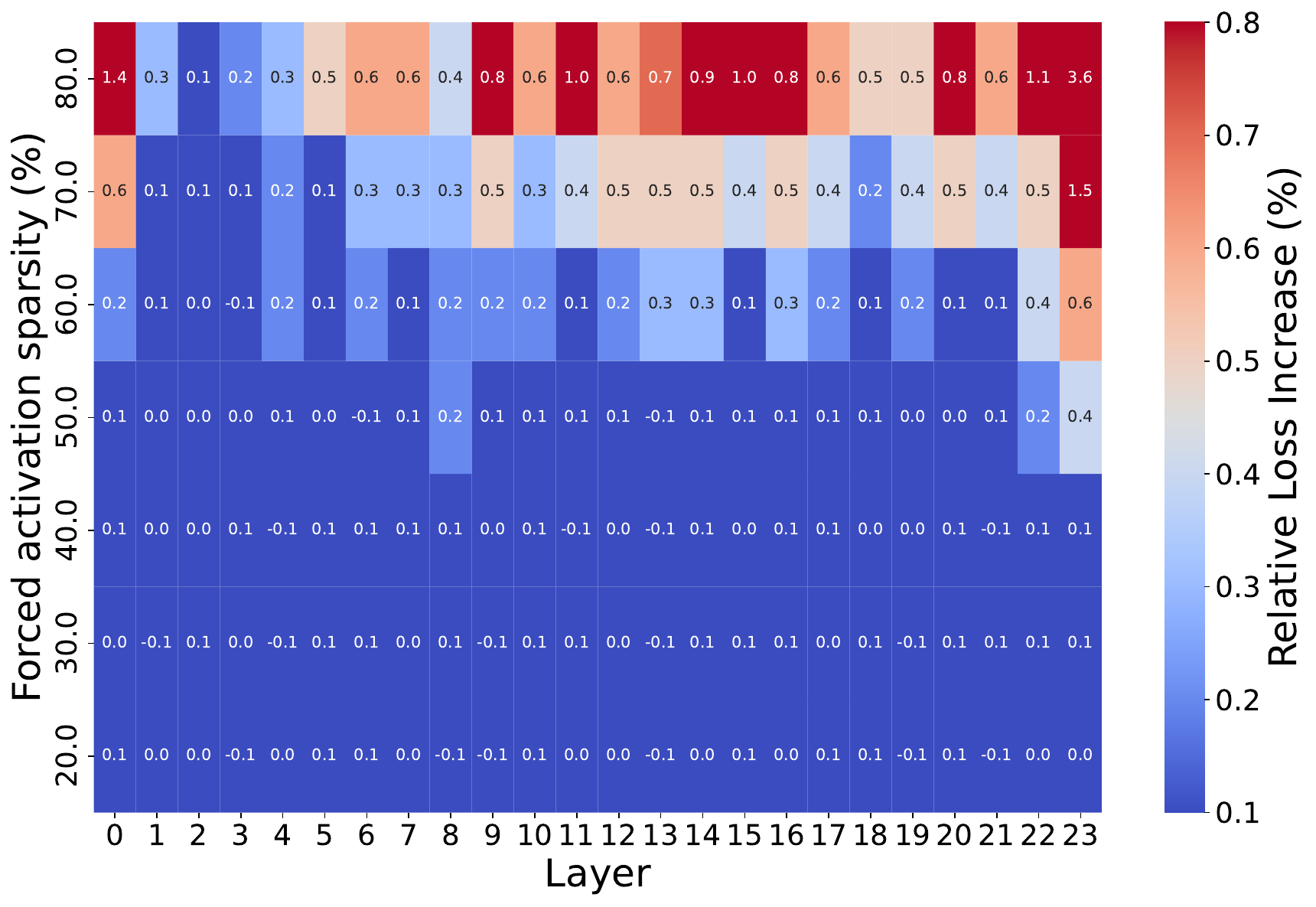}
            \caption{Layer-wise sensitivity to forced sparsity.}
            \label{fig:per_projection_sensitivity}
        \end{subfigure}
        
    \end{minipage}
    
    \caption{Sensitivity analysis to different degrees of forced top-k sparsity in the MMFreeLM model, showing the increase in loss over the dense base-model when varying levels of sparsity are enforced on a per projection type basis (left) and on per layer basis (right) .}
    \label{fig:sparsity_sensitivity}
\end{figure*}

\section{Related work}
Smoother non-saturating activations, such as GELU \cite{gelu_act} and SiLU/Swish \cite{swish_act}, have largely replaced ReLU due to improved optimization stability and downstream performance \cite{smooth_acts_survey}. Unlike ReLU, these functions do not naturally produce zero activations. Recent studies in LLMs, however, show that switching back to ReLU can be done with minimal performance loss \cite{relu_strikes_back}, with new variants like ReLU² \cite{relu2_wins} and dReLU \cite{turbosparse} that aim to restore or enhance sparsity while retaining competitive performance.

TurboSparse \cite{turbosparse} focuses on sparsity in the feed-forward network (FFN) by introducing the dReLU activation function. The Swish activation in the SwiGLU block is replaced with ReLU, and another ReLU is added to the Up projection. By continued pre-training of these modified models, they are able to recover most of the performance of the dense baseline on benchmark tasks. The sparsity they achieve, however, is localized. Only the inputs to the Down projection are zeroed, while most other projections remain dense. As a result, even though sparsities above 90\% are reported in parts of the FFN, the overall proportion of active parameters across the model is much lower. 

Other approaches have tried to achieve more model-level sparsity rather than only within the FFN. Q-Sparse \cite{qsparse_topk} does this by applying top-$K$ selection to activations and using a straight-through estimator to preserve gradients, combined with squared ReLU to promote sparsity. However, selecting the K largest magnitude activation requires sorting activations on a per-token basis, potentially introducing significant overheads, especially on constrained edge-hardware. Additionally, it requires synchronization across channels, complicating implementation on compute-memory integrated platforms \cite{pierro2025acceleratinglinearrecurrentneural}. This is particularly problematic in many-core systems where a single layer may be distributed across multiple cores: identifying the global top-$K$ requires a synchronization barrier across all participating cores before any MAC computation can proceed, introducing inter-core communication overhead that can negate the latency savings from sparsity. TEAL \cite{teal_train_free} takes a different direction by introducing a layer-wise sparsification strategy. It computes token importance scores and selectively keeps only the most relevant tokens at each layer, allowing each layer to tolerate different levels of sparsity. However, this approach is only applied for the decode phase of inference, leaving the pre-fill phase fully dense and limiting potential end-to-end efficiency improvements.

\section{Background}
\subsection{Activation Sparsity in Neural Networks}
\begin{figure}[H]
    \centering
    \includegraphics[width=\columnwidth]{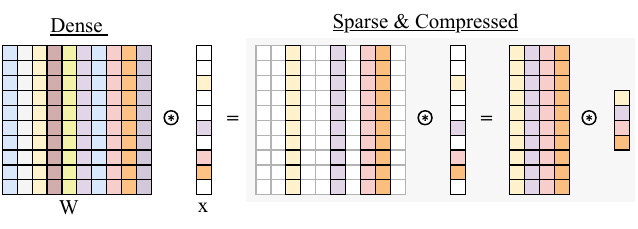}
    \caption{Illustration of activation sparsity in a matrix-vector multiplication, where zero-valued activations allow skipping associated weight accesses of entire rows in the weight matrix.}
    \label{fig:sparse_mvm}
\end{figure}

In an MVM operation, a zero-valued activation implies that all multiply-accumulate (MAC) operations involving its corresponding column of the weight matrix contribute nothing to the output. As shown in Fig.~\ref{fig:sparse_mvm}, this enables structured skipping: entire columns of weights associated with zero activations can be bypassed, eliminating both the MAC operations and the need to fetch those weights from memory. Since MVMs are often memory-bound, where performance is constrained more by the cost of moving data than by raw compute throughput, reducing memory accesses can directly yield substantial energy and latency gains \cite{eyeriss_v2}. By avoiding both the computations and memory accesses for weights corresponding to zero-valued activations, we can save bandwidth, reduce cache pressure, improve latency, and lower overall energy consumption.

If $\rho$ denotes the fraction of zero activations, then in the ideal case, both the memory reads and the number of computations scale proportionally with $(1-\rho)$. This means that the ideal throughput of hardware that can support sparsity is simply the dense throughput over the share of non-zero activations, i.e., $f_{\text{sparse}} = f_{\text{dense}}/ (1 - \rho)$. 

GPUs tend to struggle to exploit activation sparsity in inference because their architecture is built for dense, highly parallel computation with regular memory access. Unstructured sparse activations break these patterns: nonzero elements are irregularly distributed, requiring indexing and indirection that hurt coalesced memory access and reduce arithmetic efficiency. Since GPU threads execute in lockstep, skipping zeros directly is difficult without wasting compute lanes. To take advantage of sparsity, weights would need to be stored in column-major order, so that all weights linked to a nonzero activation can be fetched contiguously; however, GPUs and their libraries are optimized for dense, row-major, or block layouts, and reordering weights adds overhead. While sparse kernels can exploit activation sparsity to some extent \cite{dejavu_sparse_kernel}, real-world speedups on GPUs are often much lower due to overheads from irregular memory access of dynamic zero-activation patterns and the limited ability of standard hardware and software to take full advantage of unstructured sparsity in a non-training setting, where the available hardware cannot be saturated with massive batch sizes. 

\subsection{Leveraging activation Sparsity on Neuromorphic Hardware Accelerators}

Activation sparsity has been extensively investigated in ASIC hardware due to the potential gains over GPUs, whose more regular and highly parallelized architecture cannot fully exploit it \cite{verhelst_sparsity_study}. Accelerators such as Eyeriss v2 \cite{Chen2018_EYERISS2} and SCNN \cite{Keckler2017_SCNN} primarily target convolutional networks, leveraging sparsity to reduce power consumption and increase inference throughput. More recently, neuromorphic computing has renewed interest in hardware optimized for sparse, event-driven activity \cite{SNPU, nexus, async_snn}, although deployment of large models on the multi-million to billion-parameter scale required for language modeling remains limited in academic chips.

Loihi 2 \cite{Intel2021_LOIHI2} represents a state-of-the-art implementation of this approach, designed for sparse, event-based neural networks. By focusing on local event-driven computation, Loihi 2 efficiently leverages both sparse weight matrices and dynamically unstructured sparse activations using fixed-point arithmetic. In multi-chip setups, the system further leverages sparse activations through an event-driven inter-chip and inter-core communication, minimizing communication overhead by transmitting only non-zero packages.

\section{Methods}
\subsection{Model selection}
To meet the demands of compact models under constrained power budgets and the need for low-latency, real-time inference at the edge, various linear-attention-based quantized LLMs have been proposed \cite{abreu2025neuromorphic, chiang2025quamba2robustscalableposttraining}. The MatMul-Free Language Model (MMFreeLM) \cite{mmfree} pushes quantization to an extreme with ternary weights ($-1, 0, +1$), together with low-precision 8-bit fixed-point activations, transforming dense layers into BitLinear layers that reduce multiplications to simple additions and subtractions. Based on the Gated Recurrent Unit (GRU) proposed by \cite{GRU}, MMFreeLM uses a MMFree-Linear-GRU (MLGRU) with ternary weights, in place of the traditional self-attention mechanism in transformers. 
With these features combined, the MMFreeLM model has proven to be well-suited for energy-efficient inference across GPUs, edge-GPU, and neuromorphic hardware \cite{mmfree, abreu2025neuromorphic}. To date, it is the only billion-parameter language model deployed on neuromorphic hardware, making it the natural target for sparsification studies aimed at improving efficiency on multi-core, multi-chip neuromorphic platforms.

\subsection{Motivating study}
\label{sec:sensitivity_analysis}

When extending sparsity beyond dense FFN layers to the full model, care must be taken since different linear projections vary in their role and sensitivity to pruning/sparsification \cite{llm_sensitivity_analysis}. In SSMs, components closely tied to their linear attention, such as projections tied to the hidden state transition $h[t] \to h[t+1]$, may be especially sensitive due to their stateful nature. To study projection-wise and layer-wise sensitivity, we injected a forced sparsity into a pre-trained MMFreeLM using top-k selection of the activations with the largest magnitude, applied either per projection type (uniform across layers) or per layer (uniform across projections). Results, shown in Fig.~\ref{fig:sparsity_sensitivity}, normalize loss increases by each component’s share of FLOPs, highlighting where in the network inactive neurons provide the best tradeoff between active parameters and performance.

\textbf{Projection-wise analysis}: The projection-wise analysis shows that along with the very sensitive LM head, the projections directly involved in the next state transition ($h[t]\rightarrow h[t+1]$), $I$ and $F$, are very sensitive to enforced sparsity, whereas projections involved in the output calculation based on the current state and input $x[t]$ are less sensitive. The analysis further highlights the resilience of the large down-projection in the FFN to forced sparsity, which again aligns well with prior sparsification efforts that have targeted this projection with great success \cite{relu2_wins, turbosparse}. Overall, these findings align well with prior work focusing on weight pruning: in transformers, pruning attention heads leads to larger performance drops than pruning feed-forward layers \cite{michel2019six,voita2019analyzing}, and in SSMs such as Mamba, stateful steps are much less tolerant of high sparsity than input projections or dense layers \cite{dao2025on}. 

\textbf{Layer-wise analysis}: The layer-wise analysis, contrary to a previous sensitivity study on transformer-based models by \cite{llm_sensitivity_analysis}, shows that there are peaks in the sensitivity of the very first and very last layers, as well as the middle layers. This observation aligns with some other recent studies on transformer architectures, which suggest that middle layers often possess greater redundancy and robustness compared to the more critical early and late layers. For example, \cite{ikeda2025layerwise} conducted a layer-wise importance analysis of feed-forward networks in transformer-based language models, showing that concentrating model capacity in the middle layers while reducing or removing components in the early and late layers improves downstream task performance. 

\subsection{Proposed Sparsification Method}  

Building on the sensitivity analysis in Section~\ref{sec:sensitivity_analysis} and prior work on activation sparsity in LLMs, which showed that activations often follow Gaussian or Laplacian distributions with near-zero mean \cite{teal_train_free}, we propose a sparsity-inducing pre-activation applied to the input of every linear projection. Combined with an $L_0$ surrogate loss penalty, this mechanism encourages activations to collapse toward zero, while accounting for the varying sensitivities of different projections and layers to balance task performance with activation sparsity on a model-wide level.

\begin{figure}
    \centering
    \includegraphics[width=1\columnwidth]{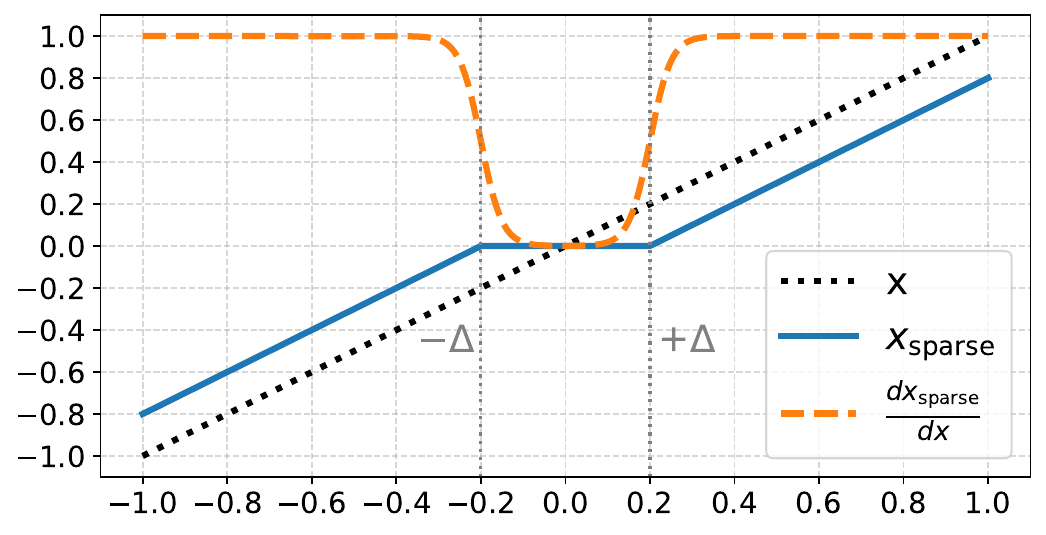}
    \caption{The sparse pre-activation function, with activations being zeroed out within the range $\pm \Delta$, along with the smoothed out surrogate gradient during the backwards pass. }
    \label{fig:sparse_mvm_eq}

\end{figure}

\textbf{Sparse per-projection pre-activation}:  
The proposed pre-activation is presented in equation \ref{eq:sparse_x} and illustrated in Fig.~\ref{fig:sparse_mvm_eq}. It consists of a two-sided ReLU that zeros out activations within the range $\pm\Delta$. This preserves the overall distribution of activations by retaining both positive and negative activations, introducing only a constant offset of $\pm\Delta$ outside the zero region. The threshold $\Delta$ is treated as a learnable parameter and optimized separately for each projection during training.

\begin{equation}
x_{\text{sparse}} = \operatorname{sign}(x) \cdot \operatorname{ReLU}(|x| - \Delta)
\label{eq:sparse_x}
\end{equation}

\begin{equation}
\frac{d x_{\text{sparse}}}{dx}_{\text{smooth}} = \sigma\big(C (|x| - \Delta)\big)
\label{eq:sparse_dx}
\end{equation}

\vspace{10pt}

Similar to findings showing the smooth nature of the SiLU activation improves learning performance, especially at high levels of sparsity when a large share of gradients would be fully zeroed out by a ReLU activation (i.e. dead neurons) \cite{deadrelu, smooth_acts_survey}, we found that a smooth surrogate for the backwards pass, described in equation \ref{eq:sparse_dx}, gave slightly better convergence with the base models' training trajectory when compared to a hard magnitude thresholding, especially at higher sparsity levels. A slope parameter $C$ controls the steepness of the smooth mask for the derivative, with larger $C$ values pushing activations toward zero more aggressively. $C$ is fixed during training; empirically we found that $C=20$ gives good results.  

\textbf{Loss penalty and differentiable sparsity surrogate}: 
To encourage the model to learn to push activations to the range within $\pm\Delta$, as well as to learn an optimal value $\Delta$ on a per-projection basis, we use an $L_0$ loss penalty added to the main task loss. Since directly counting zeros in the activation vector would obstruct gradient flow to this penalty, we instead employ a surrogate sparsity measure $\hat{s}$, resulting in the following learning objective:

\begin{equation}
\mathcal{L} = \mathcal{L}_{\text{task}} + \lambda \, (1 - \hat{s}), 
\quad 
\hat{s} = \frac{1}{N} \sum_{i=1}^N \exp\!\big(-k \, |x_{\text{q-sparse},i}|\big),
\label{eq:loss_with_sparsity}
\end{equation}

where $\mathcal{L}_{\text{task}}$ is the primary task loss (e.g., cross-entropy),  
$\lambda$ controls the strength of the sparsity penalty, $N$ is the number of activations considered, $x_{\text{q-sparse},i}$ is the $i$-th sparse activation, $k$ is the exponential steepness parameter, and $\hat{s}$ serves as a differentiable proxy for the fraction of zero activations.  Empirically, setting $k=10$ was found to provide a good trade-off between accurately estimating true sparsity and excessively large gradient norms. The sparsity penalty is weighed on a per-layer basis, depending on the resulting reduction in MAC operations due to a zero activation in that layer. Additionally, similar to previous works on neural network pruning through regularization \cite{wang2021neuralpruninggrowingregularization}, we ramp up the penalty weight term $\lambda$ slowly at the start of training, to avoid excessive sparsity before important features of the dataset have been learned. We found that a linear warm-up of 5\% of the total training steps performed well.

\begin{figure*}[t]
    \centering
    \includegraphics[width=0.85\textwidth]{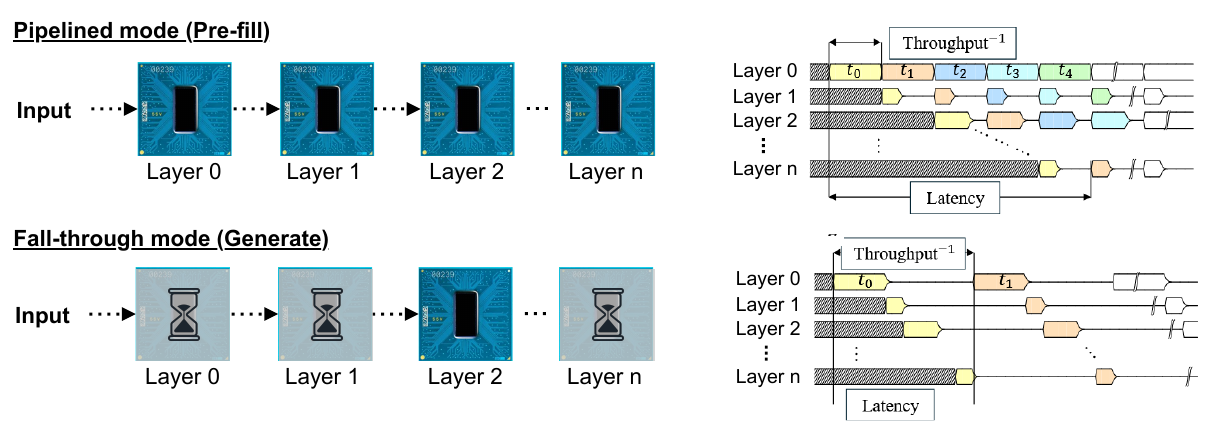}
    \caption{Different execution modes on Loihi 2, with pipelined mode (top) optimizing throughput and Fall-through mode (bottom) optimizing latency.}
    \label{fig:loihi_modes}
\end{figure*}

\subsection{Multi-Chip Deployment on Neuromorphic Hardware}

\subsubsection{Hardware Platform}
Our hardware deployment results are derived from the real-world deployment of the dense MMFreeLM model on the Loihi 2 platform. The platform supports two operating modes \cite{mmfree}, illustrated in figure \ref{fig:loihi_modes}. In pipelined mode, new inputs are introduced at every fixed time per step (TPS) and passed through successive layers, maximizing throughput. In fall-through mode, inputs are introduced only after the previous ones have been fully processed, thereby minimizing latency and allowing for a dynamically varying TPS with the per-chip workload. LLM deployment aligns naturally with these modes: prefill processing of long input sequences leverages pipelined mode for throughput efficiency, while autoregressive token generation relies on fall-through mode, since producing token $t$ must complete before token $t+1$ can be processed.

Viewed as a many-core SoC pipeline, this deployment benefits from activation sparsity at two levels of the system hierarchy: intra-chip, where zero-skipping within each core reduces MAC operations and local memory traffic, and inter-chip, where event-driven links suppress zero-valued packets to cut communication overhead. Both effects are formalized in the performance model below.

\subsection{Performance Benchmarking and Modeling}

We extend the performance modeling framework of \cite{abreu2025neuromorphic} to account for the impact of activation sparsity on throughput, latency and power. Starting from the dense baseline, we first derive the effective MAC density $r$ from the layer-level structure of each block, and then use $r$ to model how sparsity modifies multi-chip throughput under the two Loihi~2 execution modes.

\subsubsection{Dense Baseline.}
The MMFreeLM architecture consists of $N_{\text{blocks}}$ sequential computational blocks, each mapped to a separate Loihi~2 core or chip. For the 370M parameter model, $N_{\text{blocks}} = 24$, where each block comprises an MLGRU token-mixing unit and a ternary FFN channel-mixing block \cite{mmfree}. 

We adopt as baseline the \emph{measured dense multi-chip throughput} $f_{\text{dense}}^{\text{generate}}$ and $f_{\text{dense}}^{\text{prefill}}$ reported by \cite{abreu2025neuromorphic}. Results are provided for two inference modes: (i)~\emph{prefill}, where tokens are processed in a pipelined manner with all layers active concurrently, and (ii)~\emph{generate}, where tokens are produced autoregressively with one layer active at a time. Energy per token follows directly as $E_{\text{dense}} \propto 1/f_{\text{dense}}$, since Loihi~2 operates under an approximately constant power envelope \cite{abreu2025neuromorphic}. This throughput--energy pair serves as the reference for all sparse extensions.

\subsubsection{Impact of sparse activations}
\label{sec:mac_density}
We restrict the analysis to matrix-vector multiplication (MVM) operations, which dominate total FLOPs; nonlinearities and scalar operations (e.g., sigmoid, bias additions) are excluded, consistent with standard FLOP accounting \cite{flops_ignoring_acts}. The FFN in each block contains a fused Up/Gate projection of size $d_h \times 2d_i$ and a Down projection of size $d_i \times d_h$; the MLGRU block contains four projections (i, f, g, o), each of size $d_h \times d_h$.

For projection $j$, let $\rho_j \in [0,1]$ denote the activation density (fraction of nonzeros) at its input. Since each nonzero activation results in access to one row of the weight matrix, MAC counts scale linearly with $\rho_j$:
\begin{align}
    \text{MAC}_{\text{dense}}  &= \sum_{j} d_{\text{in}}^{(j)} \cdot d_{\text{out}}^{(j)},  \label{eq:mac_dense}  \\
    \text{MAC}_{\text{sparse}} &= \sum_{j} \rho_j \; d_{\text{in}}^{(j)} \cdot d_{\text{out}}^{(j)}. \label{eq:mac_sparse}
\end{align}
The effective MAC density $r$, defined as the ratio of sparse to dense MACs,
\begin{equation}
    r \;=\; \frac{\text{MAC}_{\text{sparse}}}{\text{MAC}_{\text{dense}}},
    \label{eq:mac_density_def}
\end{equation}
is therefore a \emph{size-weighted average} of the per-projection densities (not a simple mean, since projection sizes differ across layers):
\begin{equation}
    r = \frac{\displaystyle\sum_j \rho_j \cdot d_{\text{in}}^{(j)} \cdot d_{\text{out}}^{(j)}}
             {\displaystyle\sum_j d_{\text{in}}^{(j)} \cdot d_{\text{out}}^{(j)}}.
    \label{eq:mac_r}
\end{equation}

Per-block latency scales approximately linearly with $r$, consistent with prior measurements on sparse accelerators \cite{Keckler2017_SCNN, nexus}. Activation sparsity therefore modifies the dense baseline through two multiplicative speedup factors:

\textbf{(i) Inter-chip communication speedup.}
Prior measurements from deployment show that dense multi-chip inference suffers a $\approx 20\%$ throughput reduction from inter-chip communication overhead \cite{abreu2025neuromorphic}. Because Loihi 2 uses event-driven communication, where zero-event packets are skipped \cite{Intel2021_LOIHI2}, this communication overhead can be reduced by sparse packets at the block-boundaries. This principle generalizes to any multi-chip system with traffic-proportional interconnect cost: since the communication penalty scales directly with activation density, increasing model-level sparsity is equivalent to reducing the effective interconnect load. We therefore model the sparsity-dependent inter-chip penalty as  
\begin{equation}
S_{\text{comm}}(\rho_{\text{com}}) = \frac{1}{0.8 + 0.2\,\rho_{\text{com}}},
\label{eq:speedup_comm}
\end{equation}
which reduces to $1$ (no gain) when $\rho_{\text{com}} = 1$ (fully dense), recovering the $f_{\text{dense}}$ baseline of \cite{abreu2025neuromorphic}.

\textbf{(ii) MAC density speedup.}
Within each block, latency scales in proportion to MAC density $r$. The intra-block factor, introduced by the zero-skipping of MAC operations, is:
\begin{equation}
S_{\text{MAC}}(r) = \frac{1}{r}.
\label{eq:speedup_mac}
\end{equation}

\textbf{Combined throughput.}
The sparse-mode throughput is obtained by multiplying the dense baseline with both factors:  
\begin{equation}
f_{\text{sparse}} = f_{\text{dense}} \cdot S_{\text{comm}}(\rho_{\text{com}}) \cdot S_{\text{MAC}}(r).
\label{eq:sparse_throughput}
\end{equation}
Lastly, due to the different execution modes on Loihi 2 used during prefill and generate, the latency reduction modeled with $S_{\text{MAC}}$ varies slightly between the two modes:
\begin{itemize}
    \item \textbf{Prefill:} Execution is pipelined, so throughput is bottlenecked by the slowest block. Using the maximum MAC density $r_{\max}$ across blocks:
    \begin{equation}
    f_{\text{sparse}}^{\text{prefill}} = f_{\text{dense}}^{\text{prefill}} \cdot \frac{1}{0.8 + 0.2\,\rho_{\text{com}}} \cdot \frac{1}{r_{\max}}.
    \label{eq:prefill}
    \end{equation}

    \item \textbf{Generate:} Execution is sequential, so latency adds linearly across blocks. Using the mean MAC density $r_{\text{avg}}$:
    \begin{equation}
    f_{\text{sparse}}^{\text{generate}} = f_{\text{dense}}^{\text{generate}} \cdot \frac{1}{0.8 + 0.2\,\rho_{\text{com}}} \cdot \frac{1}{r_{\text{avg}}}.
    \label{eq:generate}
    \end{equation}
\end{itemize}

\section{Results}

\begin{table*}[t]
\centering
\small
\caption{Throughput and efficiency across of various dense and sparse language models, including our sparse MMFreeLM, for prefill and generation across various sequence lengths, running on a NVIDIA H100 GPU, Intel's Loihi 2 and a Nvidia Jetson.}
\label{tab:loihi_results}
\vspace{0.25em}

\begin{tabular}{@{}l l l *{4}{c} *{4}{c}@{}}
\toprule
 & & & \multicolumn{4}{c}{Throughput ($\uparrow$ tokens/sec)} & \multicolumn{4}{c}{Efficiency ($\downarrow$ mJ/token)} \\
\cmidrule(lr){4-7}\cmidrule(lr){8-11}
\multicolumn{3}{@{}l}{\textbf{Sequence length}} & 500 & 1000 & 4000 & 8000 & 500 & 1000 & 4000 & 8000 \\
\midrule

\multirow{6}{*}{\rotatebox{90}{\textbf{Generate}}}
  & \textbf{MMF (sparse)} & \textbf{Loihi 2\textsuperscript{\dag}} 
      & \textbf{224.1} & \textbf{224.1} & \textbf{224.1} & \textbf{224.1} 
      & \textbf{75.0} & \textbf{75.0} & \textbf{75.0} & \textbf{75.0} \\

  & MMF (dense)         & Loihi 2\textsuperscript{*}         
      & 41.5 & 41.5 & 41.5 & 41.5 
      & 405 & 405 & 405 & 405 \\
  & MMF (dense)         & H100 \textsuperscript{\ddag}                                        
      & 13.4 & 13.3 & 13.5 & 13.2 
      & 10.1k & 10.1k & 10.0k & 9.9k \\
  & TF++          & H100 \textsuperscript{\ddag}                                        
      & 22.4 & 22.9 & 21.7 & 21.3 
      & 5.5k & 5.6k & 6.2k & 6.8k \\
  & Alireo (400M)        & Jetson \textsuperscript{\ddag}                                 
      & 14.3 & 14.9 & 14.7 & 15.2 
      & 723 & 719 & 853 & 812 \\
  & Qwen2 (500M)         & Jetson \textsuperscript{\ddag}                   
      & 13.4 & 14.0 & 14.1 & 15.4 
      & 791 & 785 & 912 & 839 \\
\midrule

\multirow{6}{*}{\rotatebox{90}{\textbf{Prefill}}}
  & \textbf{MMF(sparse)} & \textbf{Loihi 2\textsuperscript{\dag}} 
      & \textbf{23.2k} & \textbf{23.2k} & \textbf{23.2k} & \textbf{23.2k} 
      & \textbf{1.1} & \textbf{1.1} & \textbf{1.1} & \textbf{1.1} \\
  & MMF (dense) & Loihi 2\textsuperscript{*}         
      & 6632 & 6632 & 6632 & 6632 
      & 3.7 & 3.7 & 3.7 & 3.7 \\
  & MMF (dense) & H100 \textsuperscript{\ddag}                                        
      & 11.4k & 13.1k & 30.6k & 51.6k 
      & 6.1 & 5.3 & 2.5 & 1.4 \\
  & TF++          & H100 \textsuperscript{\ddag}                                       
      & 21.6k & 32.7k & 44.3k & 55.4k 
      & 11.3 & 7.3 & 5.4 & 4.3 \\
  & Alireo (400M)        & Jetson \textsuperscript{\ddag}                                 
      & 849  & 1620 & 3153 & 2258 
      & 11.7 & 7.8 & 6.8 & 7.6 \\
  & Qwen2 (500M)         & Jetson \textsuperscript{\ddag}                                 
      & 627  & 909  & 2639 & 3861 
      & 17.9 & 13.9 & 6.7 & 4.4 \\
\bottomrule

\end{tabular}
\begin{minipage}{0.98\linewidth}
\vspace{0.35em}
\footnotesize
\textsuperscript{\dag} Proposed sparse model ($\lambda = 2.0$) with metrics extrapolated from dense  model using equations \ref{eq:prefill} and \ref{eq:generate} \\
\textsuperscript{*} Baseline deployment results of 370M dense MMFreeLM in multi-chip setup from \cite{abreu2025neuromorphic}. Includes inter-chip communication slowdown over single-chip measurements. \\
\textsuperscript{\ddag} Jetson and H100 metrics from reported deployment by \cite{mmfree}.


\end{minipage}
\label{tab:mmfree_style_full}

\end{table*}

\subsection{Training setup}
We continue training pre-trained 370M and 2.7B MMFreeLM models on 4B tokens of the FineWebEdu dataset \cite{lozhkov2024fineweb-edu} using a cosine learning rate schedule with a reduced initial rate to preserve patterns from the original training. For comparison, we include ReLU-fication \cite{relu_strikes_back} and the dReLU method \cite{turbosparse} baselines, as well as a continued-training baseline to control for the effect of the additional training data. We train three models with our proposed method at varying sparsity penalty strengths.

\subsection{Sparsity of trained models}

We present the effective MAC of the evaluated sparsification methods compared to our proposed method, along with a per-projection type sparsity, divided into the FFN and MLGRU blocks, in table \ref{tab:sparsity_results}. Activation sparsities are captured over the entire benchmark set used in section \ref{sec:reasoning_benchmarks}.

\begin{table}[H]
\centering
\footnotesize
\caption{Activation sparsity (\%) per projection for different sparsification methods on MMFreeLM.}
\label{tab:sparsity_results}

\setlength{\tabcolsep}{1.4pt}
\renewcommand{\arraystretch}{0.85}

\begin{tabular}{lccccccccc}
\toprule
& \multicolumn{4}{c}{MLGRU} & \multicolumn{2}{c}{FFN} & Head & \\
\cmidrule(lr){2-5} \cmidrule(lr){6-7} \cmidrule(lr){8-8}
Method  & I & F & G & O & Up & Down & & MAC sparsity $(\uparrow)$\textsuperscript{\dag} \\
\midrule
\multicolumn{9}{c}{370M MMFreeLM} \\
\midrule
Baseline (SiLU)           & 1.1  & 3.2  & 1.5  & 37.3 & 1.1  & 30.1 & 2.2  & 10.0\textsuperscript{*} (10.9) \\
ReLU                      & 1.1  & 2.8  & 1.4  & 37.9 & 1.2  & 87.6 & 2.2  & 23.7\textsuperscript{*} (21.6) \\
dReLU                     & 1.1  & 2.8  & 1.4  & 25.0 & 1.3  & 93.2 & 2.4  & 25.1\textsuperscript{*} (22.9) \\
\textbf{Ours} ($\lambda=1.0$) & 45.7 & 61.2 & 48.9 & 78.2 & 63.8 & 92.2 & 24.0 & \textbf{69.5\textsuperscript{*} (64.3)} \\
\textbf{Ours} ($\lambda=2.0$) & 61.9 & 75.6 & 66.0 & 85.2 & 79.1 & 93.4 & 40.2 & \textbf{80.1\textsuperscript{*} (76.2)} \\
\midrule
\multicolumn{9}{c}{2.7B MMFreeLM} \\
\midrule
Baseline (SiLU)           & 1.7  & 4.3  & 1.4  & 41.8 & 1.3  & 30.4 & 2.9  & 12.5 (12.1) \\
\textbf{Ours} ($\lambda=1$)   & 45.7 & 61.1 & 40.6 & 66.3 & 56.6 & 91.0 & 9.2  & \textbf{63.2\textsuperscript{*} (61.5)} \\
\bottomrule
\end{tabular}
\begin{minipage}{0.98\linewidth}
\vspace{0.25em}
\footnotesize
\textsuperscript{*}\,Excluding LM Head, which is not included in the Loihi 2 implementation in \cite{mmfree}. Value in parentheses shows the effective MAC sparsity with the LM Head included. \\
\textsuperscript{\dag} Calculated as 1 - $r$ as described in section \ref{sec:mac_density}.  
\end{minipage}
\end{table}

Due to some inherent sparsity from the fixed-point 8-bit quantization, even the base model exhibits an average baseline parameter sparsity of 10.0\%. While both ReLU-based methods achieve significant levels of sparsity in the FFN at the input of the large down-projection, the impact on the overall share of active parameters is limited, as the FFN only makes up $\approx 2/3$ of total FLOPS, with the down-projection contributing to just $\approx1/3$ of that. The result also shows that the second ReLU activation inserted with the dReLU method has a limited impact on model-wide sparsity, as the dot-product between the sparse post-activation output of the gate projection with the dense Up projection in and of itself already results in a sparse vector. Our proposed method is able to achieve  significantly higher levels of model-wide sparsity by not only targeting linear projections where SiLU activations can be replaced by ReLU, but all linear projections in the model.


\subsection{Performance on Reasoning Tasks}
\label{sec:reasoning_benchmarks}

\begin{table}[H]
\centering
\small
\caption{Zero-shot accuracy of sparse MMFreeLM models compared to the dense baseline. All models use ternary weights and 8-bit activations.}
\label{tab:reasoning_benchmark}

\setlength{\tabcolsep}{3.5pt}
\renewcommand{\arraystretch}{0.92}

\begin{tabular}{lcccccc}
\toprule
Model & MACs ($\downarrow$) & ARCe & HS & OQ & PQ & Avg \\
\midrule
\multicolumn{7}{c}{370M MMFreeLM} \\
\midrule
Baseline (SiLU) & 307M & 41.54 & 32.69 & 30.40 & 62.89 & 41.88 \\
ReLU            & 267M & 39.90 & 32.99 & 31.40 & 61.70 & 41.50 \\
dReLU           & 263M & 38.68 & 32.08 & 29.20 & 61.04 & 40.25 \\
Ours ($\lambda=1$) & 118M & 41.29 & 31.58 & 31.00 & 61.10 & 41.24 \\
Ours ($\lambda=2$) & 95M  & 38.38 & 30.60 & 29.80 & 60.17 & 39.74 \\
\midrule
\multicolumn{7}{c}{2.7B MMFreeLM} \\
\midrule
Baseline (SiLU)    & 2.32B & 50.55 & 47.54 & 35.00 & 69.26 & 50.59 \\
Ours ($\lambda=1$) & 1.01B & 48.32 & 43.43 & 35.80 & 66.43 & 48.50 \\
\bottomrule
\end{tabular}

\vspace{3pt}
\begin{minipage}{0.98\linewidth}
\end{minipage}

\end{table}

We evaluated the zero-shot performance of the sparsified models on the same set of language tasks as in the original MMFreeLM work, including ARC-Easy, ARC-Challenge~\cite{clark2018think,yadav2019quick}, HellaSwag~\cite{zellers2019hellaswag}, OpenBookQA~\cite{mihaylov2018can}, PIQA~\cite{bisk2020piqa}, and WinoGrande~\cite{sakaguchi2020winogrande}. The results are presented in table \ref{tab:reasoning_benchmark}, showing a small degradation in the average reasoning task performance compared to the dense baseline model, with the 370M ($\lambda=1$) model outperforming the dReLU model at just half the average active MAC operations.


\subsection{Energy efficiency of sparse model}

We apply the methodology described in \ref{sec:sensitivity_analysis} to the 24-chip Loihi measurements in \cite{abreu2025neuromorphic} on our sparse 370M-parameter model ($\lambda$ = 2) to estimate the performance gains of a sparse model. Our model shows a worst block MAC density of $r_{max}=0.31$ in layer 17. The input activation density of the same block is taken as the average activation sparsity to the MLGRU (i, f \& g-proj) of the same layer (see \cite{mmfree} for mapping details) and calculated to $\rho_{com}=0.61$. Using equation \ref{eq:prefill}, we calculate a decrease in latency and energy-per-token of 3.5$\times$ against the dense deployment for prefill. 

For generate, we use equation \ref{eq:generate} with an average model-wide MAC density of $r_{avg} = 0.20$ and an average $\rho_{com}=0.67$ to calculate an improvement in both metrics of 5.4$\times$ as compared to the dense baseline. 

For further comparison, we also include deployment metrics by \cite{mmfree} of transformer models with comparable downstream task performance, including the 500M parameter Qwen2 model \cite{yang2024qwen2technicalreport}, and a 400M parameter Alireo model \cite{aliero} on GPU and edge-GPU (Jetson).


\section{Conclusion}

We present a method for inducing high activation sparsity in heavily quantized linear-attention models through learnable sparsifying pre-activations, achieving up to 76\% reduction in MAC operations with minimal performance loss. Sparsity reduces system cost at two levels in a multi-chip deployment: intra-chip, by skipping zero-activation MAC operations within each processing element, and inter-chip, by reducing activation volume transmitted across the chip network. Applying our performance model to the only billion-parameter model deployed on a neuromorphic multi-chip platform, we project up to 5.4$\times$ gains in throughput and energy efficiency over the dense baseline, and up to 37$\times$ over a comparable transformer on edge GPU, gains that GPU architectures, constrained by their regular memory hierarchies, cannot realize from unstructured sparsity alone. These results demonstrate that combining dynamic activation sparsity with aggressively quantized recurrent models is a natural fit for event-driven multi-core, multi-chip platforms.

\bibliographystyle{IEEEtran}
\bibliography{sample-base}

\end{document}